\documentclass{article}
\usepackage{spconf,amsmath,epsfig}
\usepackage{balance}
\usepackage{cite}
\usepackage{graphicx}

\usepackage{lipsum}
\usepackage{url}
\usepackage{caption}
\usepackage{subfig} %should come after caption package if given, can't be used with caption2, defines subfloat
\usepackage{color}
\usepackage{balance}
\usepackage{float}
\usepackage{multirow}
\usepackage{float}
\usepackage{breqn}

\begin{document}\sloppy

% Example definitions.
% --------------------
\def\x{{\mathbf x}}
\def\L{{\cal L}}

\title{Localizing Adverts in Outdoor Scenes}

\makeatletter
\def\@name{
\emph{Soumyabrata Dev$^{1}$, Murhaf Hossari$^{1}$, Matthew Nicholson$^{1}$, Killian McCabe$^{1}$, Atul Nautiyal$^{1}$},\\ 
\emph{Clare Conran$^{1}$, Jian Tang$^{3}$, Wei Xu$^{3}$, and Fran\c{c}ois Piti\'e$^{1,2}$}
\makeatother
\thanks{The ADAPT Centre for Digital Content Technology is funded under the SFI Research Centres Programme (Grant 13/RC/2106) and is co-funded under the European Regional Development Fund.}
\thanks{Send correspondence to F.\ Piti\'e (PITIEF@tcd.ie).}\vspace{5px}}
\address{
$^{1}$The ADAPT SFI Research Centre, Trinity College Dublin\\
$^{2}$Department of Electronic \& Electrical Engineering, Trinity College Dublin\\
$^{3}$Huawei Ireland Research Center, Dublin
}

\maketitle

\begin{abstract}
Online videos have witnessed an unprecedented growth over the last decade, owing to wide range of content creation. This provides the advertisement and marketing agencies plethora of opportunities for targeted advertisements. Such techniques involve replacing an existing advertisement in a video frame, with a new advertisement. However, such post-processing of online videos is mostly done manually by video editors. This is cumbersome and time-consuming. In this paper, we propose \emph{DeepAds} -- a deep neural network, based on the simple encoder-decoder architecture, that can accurately localize the position of an advert in a video frame. Our approach of localizing billboards in outdoor scenes using neural nets, is the first of its kind, and achieves the best performance. We benchmark our proposed method with other semantic segmentation algorithms, on a public dataset of outdoor scenes with manually annotated billboard binary maps. 
\end{abstract}
\begin{keywords}
advertisement, online videos, DeepAds.
  \vspace{-0.4cm}
\end{keywords}
\section{Introduction}
\label{sec:intro}

With the recent growth in the number of online videos, there is a massive opportunity for advertisement and marketing agencies for targeted advertisements. The consumers have an insatiable appetite for media, and impressive technological advances in the domain of multimedia is creating several avenues for the consumers to view the videos at ease. 

Imagine a video content is generated in California, United States, and it needs to be witnessed by viewers in Tokyo, Japan. The demographics and the consumer habits in California and Tokyo are considerably different. In order to provide an impactful viewing experience to viewers in Tokyo, the adverts in the online video needs to be carefully curated. It is therefore, of huge economic merit, to create an advert creation system~\cite{nautiyal2018advert}, that can accurately detect~\cite{hossari2018adnet}, and seamlessly integrate new adverts in existing on-demand video content.

Most of the existing works in the literature~\cite{covell2006advertisement,hussain2017automatic} involve identification of logos in video frames, or detecting advertisement clips in video streams. None of the work involve the general task of localizing existing billboards/advertisement boards in outdoor scenes. Recently, in ~\cite{dev2019alos}, we proposed the first large-scale dataset of outdoor scenes with manually annotated binary maps of billboard location. In this paper, our key idea is to exploit deep neural networks based segmentation algorithms on this dataset of outdoor scenes. Such approach can automatically localize the position of advertisement in a video frame, and thus assist video editors in significantly reducing the amount of manual annotation of the frames in the original video for personalized marketing. In this paper, we focus our attention on outdoor-scene videos containing street billboards. We interchangeably use the term \emph{advert} and \emph{billboard} to indicate an advertisement candidate in the video frame. 

The main contributions of this paper are: (a) we compare the performance of existing semantic segmentation deep neural network architectures for automatic billboard localization, and (b) we also propose a lightweight convolution-deconvolution architecture that achieves best-in-class performance in localizing billboard in outdoor scenes.

\vspace{-0.4cm}
\section{Billboard Localization}

\subsection{Deep-learning based architectures for semantic segmentation}
In this section, we explore the various state-of-the-art algorithms in semantic segmentation, that can be exploited to localize billboards in outdoor scene images. Prior to the release of our dataset in ~\cite{dev2019alos}, there exists no publicly available dataset of street-view images, with manually annotated billboard maps. Therefore, it is interesting for us to explore how the various recent algorithms on object instance segmentation, perform in this specific application of localizing billboards. With the success of convolutional neural networks in visual computing, several neural models were developed for the task of semantic segmentation. To the best of our knowledge, no models were developed specifically for the task of localizing billboards in outdoor scenes, for the task of targeted advertisements. 

In ~\cite{long2015fully}, Long et al.\ proposed Fully Convolutional Network (FCN) that uses convolutional layers, without the use of fully connected layers. Another popular model is U-Net~\cite{ronneberger2015u}, that uses an encoder-decoder architecture, for the task of accurate segmentation of neuronal structures in the field of bio-medical image processing.  Recently, Zhao et al.\ in ~\cite{zhao2017pyramid}, proposed a pyramid scene parsing network called PSPNet that produced impressive results in scene parsing of cityscapes dataset~\cite{cordts2016cityscapes}. More discussion on the benchmarking of these techniques on our specific task of billboard localization can be found in Section~\ref{sec:obj-eval}.

\vspace{-0.5cm}
\subsection{Proposed DeepAds model}

Most of the current deep learning techniques require a huge amount of labelled data, and extensive graphics processing unit (GPU) computation during training. Also, in a few cases, the models do not converge well. In this section, we introduce a scaled down architecture called DeepAds, that is specifically design for the task of billboard localization in street view images.

\begin{figure}[htb]
  \begin{center}
    \includegraphics[width=0.4\textwidth]{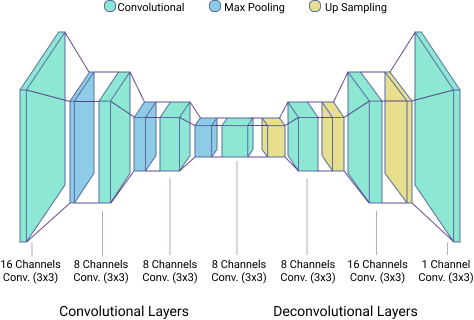}
  \end{center}
    \vspace{-0.5cm}
  \caption{Our proposed DeepAds model is based on a simple encoder-decoder architecture. The convolutional, max pooling and upsampling layers are color coded for ease of interpretation.}

  \label{fig:deepads-arch}
  \vspace{-0.5cm}
\end{figure}

Our architecture is inspired from FCN~\cite{long2015fully}, and is based on a simple encoder-decoder based architecture for the purpose of billboard localization in an image. We use a fixed input image size of $200\times200$ for training our architecture. The encoder block in DeepAds gradually reduces the spatial dimension of the input image, whereas the decoder block attempts to recover the spatial information in the image. We use convolutional filters with dimension $3\times3$ with a stride of unity. Figure~\ref{fig:deepads-arch} describes the architecture of our proposed DeepAds model. 

We use the regular RGB image of dimension $200\times200$ as the input image in DeepAds model. The output of our model is a probabilistic mask of dimension $200\times200$, wherein each pixel is assigned the probability of belonging to \emph{billboard} or \emph{no billboard} class. We use a thresholding approach to convert the probabilistic mask into a binary mask -- more details on the choice of threshold is discussed later in Section~\ref{sec:which-value}.

\section{Experiments and Results}

In \cite{dev2019alos}, we propose and release a publicly available dataset containing outdoor scene images, with high-quality annotation maps of billboard location. All the images in the dataset are manually annotated with the accurate position of the billboard in the image. The images in this dataset are mainly sourced from Mapillary -- a crowd-sourcing service for sharing geotagged photos. We benchmark our proposed model on this large-scale dataset of billboard images. Our dataset consists of a total of $9315$ images of outdoor scenes. The minimum resolution of each of the input image is $800\times600$ pixels.

\begin{figure*}[htb]
\centering
\includegraphics[height=0.135\textwidth]{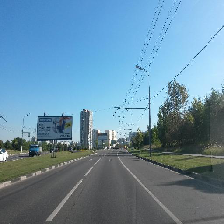}
\includegraphics[height=0.135\textwidth]{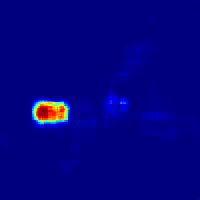}
\includegraphics[height=0.135\textwidth]{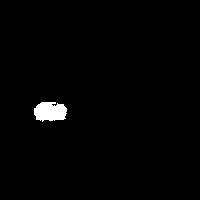}
\includegraphics[height=0.135\textwidth]{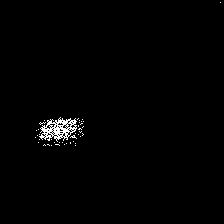}
\includegraphics[height=0.135\textwidth]{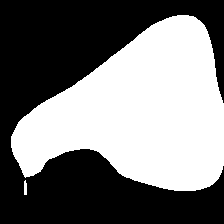}
\includegraphics[height=0.135\textwidth]{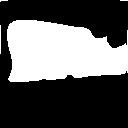}\\
\includegraphics[height=0.135\textwidth]{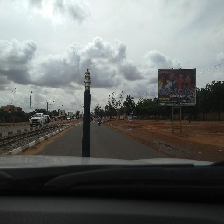}
\includegraphics[height=0.135\textwidth]{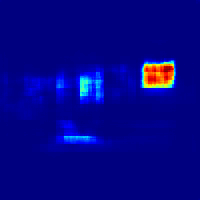}
\includegraphics[height=0.135\textwidth]{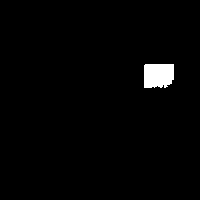}
\includegraphics[height=0.135\textwidth]{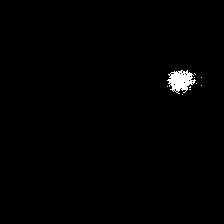}
\includegraphics[height=0.135\textwidth]{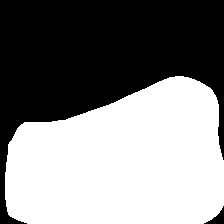}
\includegraphics[height=0.135\textwidth]{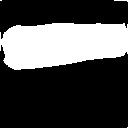}\\
\vspace{-0.3cm}
\subfloat[Input image]{\includegraphics[height=0.135\textwidth]{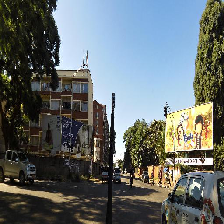}}\ 
\subfloat[Prob.\ DeepAds]{\includegraphics[height=0.135\textwidth]{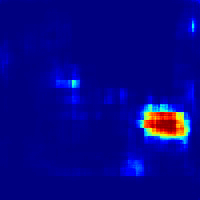}}\ 
\subfloat[Binary DeepAds]{\includegraphics[height=0.135\textwidth]{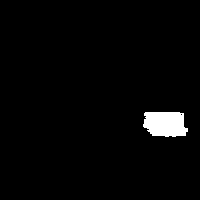}}\ 
\subfloat[FCN result]{\includegraphics[height=0.135\textwidth]{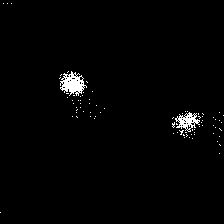}}\ 
\subfloat[PSPNet result]{\includegraphics[height=0.135\textwidth]{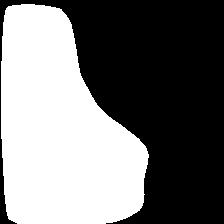}}\ 
\subfloat[U-Net result]{\includegraphics[height=0.135\textwidth]{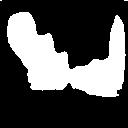}}
\caption{Subjective evaluation of several benchmarking algorithms. We represent sample (a) input images, along with (b) probabilistic output map of DeepAds model, and (c) binary output map of proposed DeepAds model. We also show the results obtained from (d) FCN, (e) PSPNet and (f) U-net for benchmarking purposes.}
\label{fig:subj-eval}
\vspace{-0.5cm}
\end{figure*}

\subsection{Subjective Evaluation}

Figure~\ref{fig:subj-eval} shows a few visual results of all the benchmarking algorithms, along with our proposed model. The probabilistic output of DeepAds provide us interesting insights -- it provides us a level of confidence for each pixel to belong to the billboard category. We illustrate the probabilistic output of our DeepAds model in Fig.~\ref{fig:subj-eval}(b). We observe that, in most cases, the probabilistic mask clearly defines the outline of the localized billboard, except for scenes that are cluttered with similar background images. We employ a simple thresholding approach with a threshold value of $0.5$ (c.f.\ Section~\ref{sec:which-value}), to convert the probabilistic map into binary map. We show the binary maps of our DeepAds model in Fig.~\ref{fig:subj-eval}(c). The results from FCN model are illustrated in Fig.~\ref{fig:subj-eval}(d). The results from FCN are coarse in nature, however, it can successfully identify the location of the billboard in the image. The binary maps computed using the PSPNet model is illustrated in Fig.~\ref{fig:subj-eval}(e). One of the disadvantage of PSPnet is the slow training, owing to small batch size during the training process. It can be further improved by synchronizing batch normalization across multiple GPUs. Finally, we show the results obtained from U-Net in Fig.~\ref{fig:subj-eval}(f). The U-Net produces a lot of false positives, in addition to identifying the region in the image containing the billboard.

\subsection{Objective Evaluation}
\label{sec:obj-eval}
In addition to the subjective evaluation, we also provide an objective evaluation of our proposed model, alongside the recent state-of-the-art algorithms. We report the average values of the following four metrics, which are commonly used in the area of semantic segmentation -- pixel accuracy, mean accuracy, mean intersection over union, and frequency-weighted intersection over union. These metrics are variations of the accuracy of pixel classification and the region intersection over union (IOU). Let us assume that $n_{ij}$ is the number of pixels in class $i$, which are predicted to belong to class $j$. The total number of classes in the task of semantic segmentation is denoted by $n_{cl}$. We define the total number of pixels in class $i$ by $t_i = \sum_{j=1}^{n_{cl}} n_{ij}$. Using these notations, we define the various metrics as follows: $\mbox{Pixel } \mbox{Accuracy } \mbox{(PA)} = \frac{\sum_{i}^{} n_{ii}}{\sum_{i}^{} t_{i}}$, $\mbox{Mean } \mbox{Accuracy } \mbox{(MA)} = \frac{1}{n_{cl}}\sum_{i}^{}\frac{n_{ii}}{t_i}$, $\mbox{Mean } \mbox{Intersection } \mbox{Over } \mbox{Union } \mbox{(mIOU)} = \frac{1}{n_{cl}}\frac{\sum_{i}^{}n_{ii}}{t_i+\sum_{j}^{}n_{ji}-n_{ii}}$, and $\mbox{Frequency } \mbox{Weighted }\mbox{Intersection } \mbox{Over } \mbox{Union } \mbox{(fwIOU)} = \frac{1}{\sum_{k}^{}t_k}\frac{\sum_{i}^{}t_in_{ii}}{t_i+\sum_{j}^{}n_{ji}-n_{ii}}$.

We compute these metrics for individual images in the testing set, using the binary predicted mask and binary ground truth map. We compute the average values of these metrics across all the testing images. Table~\ref{table:result} summarizes the results of the various benchmarking approaches in the dataset of billboard images. 

\begin{table}[htb]
%\small 
\centering
\begin{tabular}{l|c|c|c|c}
       & \textbf{PA} & \textbf{MA} & \textbf{mIOU} & \textbf{fwIOU} \\ \hline
FCN    & 0.962  & 0.699  &   0.638   &  0.937 \\ \hline
PSPNet & 0.554 & 0.558 & 0.304 & 0.521 \\ \hline
U-Net &  0.721  &  \textbf{0.814}  &  0.432 & 0.689 \\ \hline 
DeepAds & \textbf{0.971} & 0.776 & \textbf{0.722} &  \textbf{0.950} \\ 
\end{tabular}
\caption{Performance evaluation of the proposed DeepAds model in localizing adverts. The best performance in each category is marked in bold. Our proposed approach performs the best, as compared to the other benchmarking algorithms.}
\label{table:result}
\vspace{-0.2cm}
\end{table}

We observe that the PSPnet performs worse as compared to other approaches, owing to non-convergence to the optimal solution. The U-net architecture, based on encoder-decoder architecture has the best mean accuracy, but, fairs poorly on the other metrics. Our simple architecture of DeepAds can efficiently learn the semantics of the advertisements in the outdoor scenes, and provides the best scores in pixel accuracy, mean IOU, and frequency weighted IOU. Furthermore, being a simple and lightweight model, the results from the FCN network are closer to the performance of DeepAds. 

\vspace{-0.4cm}
\subsection{Determination of threshold}

\label{sec:which-value}
The output of our DeepAds model is a probabilistic image, wherein each pixel in the image is classified as either belonging to \emph{billboard} or \emph{no billboard} classes. We employ a simple thresholding approach to convert the probabilistic mask into a binary mask. We perform this thresholding experiment for varying range of threshold values in the interval $[0,1]$ with steps of $0.05$. We compute the pixel accuracy for all the testing images, across the different threshold values. The pixel classification accuracy is defined by: $\mbox{Accuracy} = \frac{TP+TN}{TP+TN+FP+FN}$, where $TP$, $TN$, $FP$ and $FN$ denote the true positives, true negatives, false positives and false negatives respectively for the binary classification task.

\begin{figure}[htb]
 \begin{center}
    \includegraphics[width=0.34\textwidth]{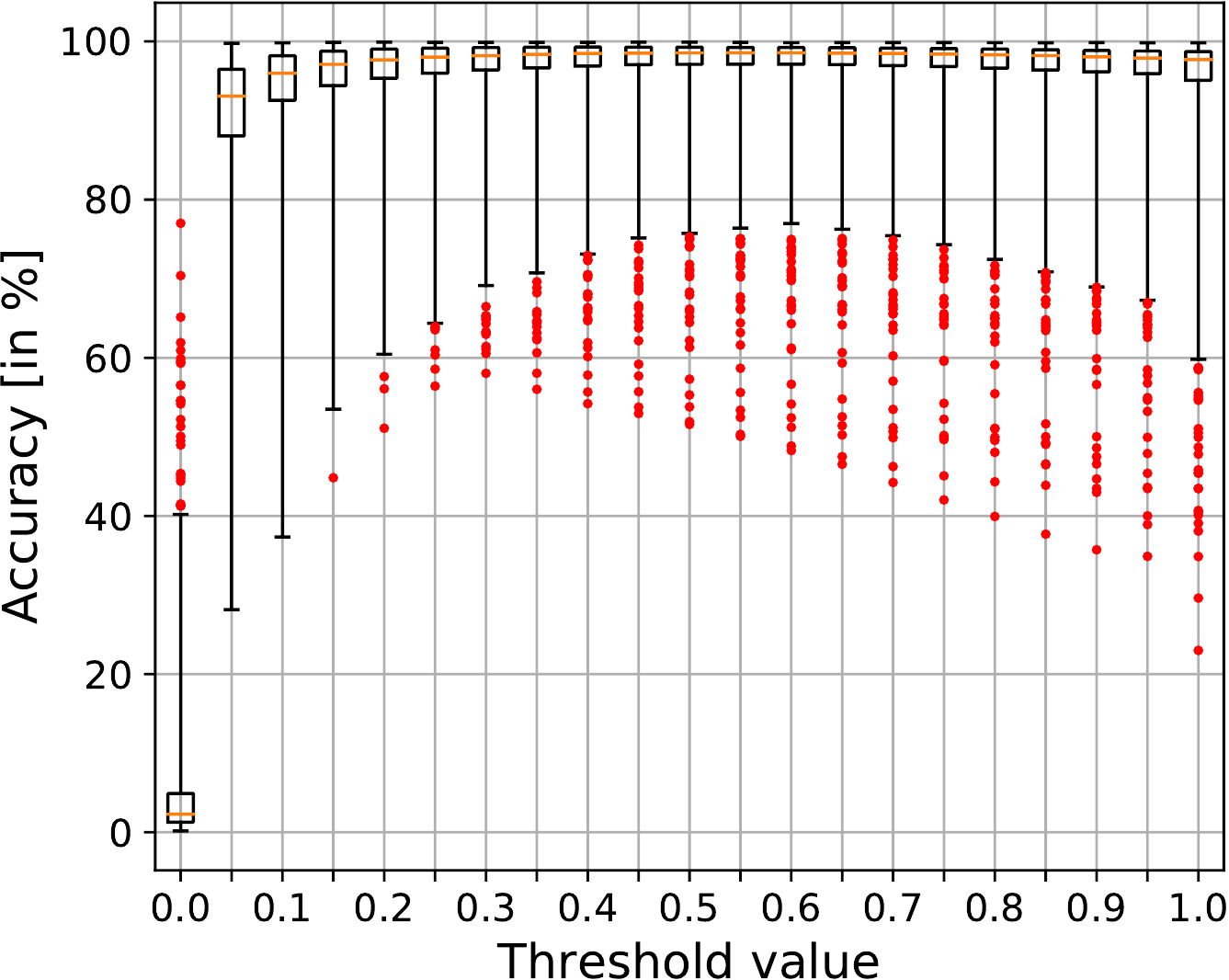}
  \end{center}
  \vspace{-0.5cm}
  \caption{Distribution of the accuracy values (measured in \%) across the different threshold values. }
  \vspace{-0.5cm}
  \label{fig:diff-th}
\end{figure}

Figure~\ref{fig:diff-th} shows the distribution of the accuracy values across all the threshold values. It is clear that the mean accuracy is low, for small- and large- values of the discrimination threshold. The largest mean accuracy corresponds for the threshold value of $0.5$ -- we chose this as the threshold to convert the probabilistic map into the binary map for evaluation purposes. Of course, this is a design parameter, and it can be further tuned based on the considered application. 

Furthermore, we also compute the receiver operating characteristics (ROC) curve for the proposed DeepAds model. The ROC curve is a plot between false positive rate (FPR) and true positive rate (TPR) for the varying threshold values in the range $[0,1]$. The FPR and TPR are defined as follows: $\mbox{FPR} = \frac{FP}{FP+TN}$, and $\mbox{TPR} = \frac{TP}{TP+FN}$.

\begin{figure}[htb]
  \begin{center}
    \includegraphics[width=0.35\textwidth]{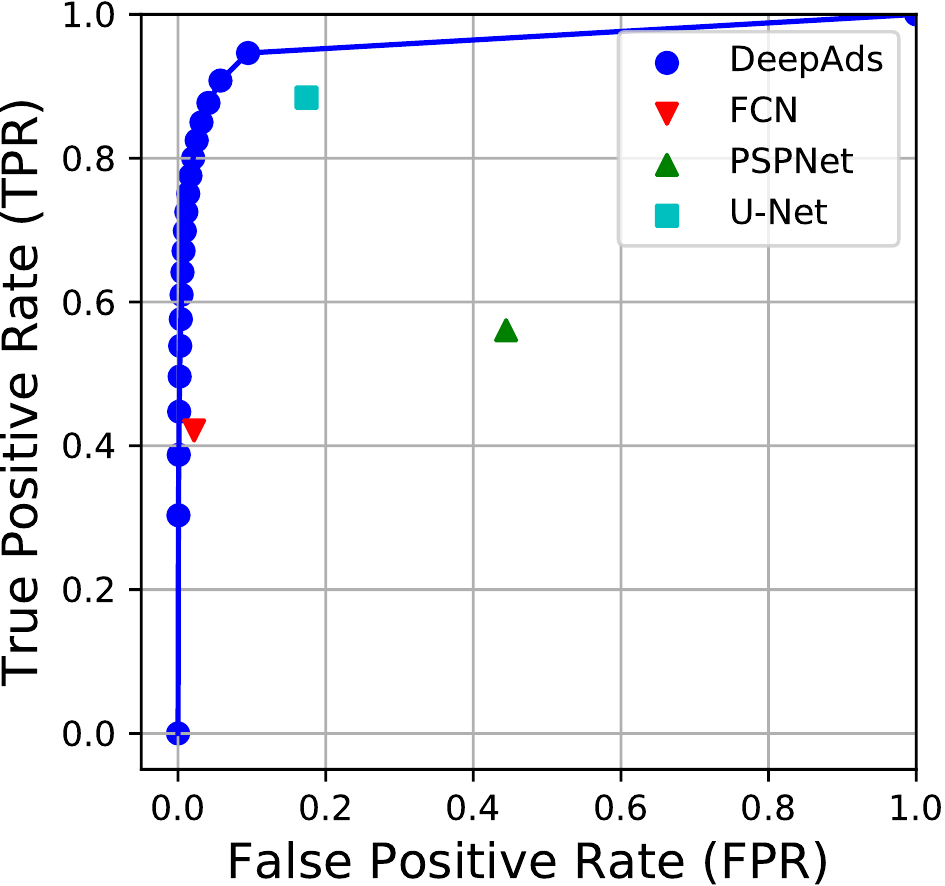}
  \end{center}
  \vspace{-0.5cm}
  \caption{Receiver operating characteristic (ROC) curve for varying thresholds in DeepAds model. We also plot the FPR and TPR values for the other benchmarking methods.}
  \label{fig:roc}
  \vspace{-0.5cm}
\end{figure}

We show the ROC curve in Fig.~\ref{fig:roc}. We also plot the corresponding values of FPR and TPR for the other benchmarking methods. As there are no thresholds involved in the other approaches, each deep neural net model correspond to a single point in the ROC plot area. We observe that our proposed DeepAds model has better performance than the other benchmarking methods, for most of the threshold values. As FCN is also a light-weight deep neural network, its performance is closer to that of DeepAds. In conclusion, we also arrive at similar conclusions from the ROC plot-- the DeepAds model with a threshold value of $0.5$, has the most competitive performance in localizing billboards in outdoor scene images, as compared to the other benchmarking methods. 

\section{Conclusion and Future Work}
\vspace{-0.3cm}
In this paper, we have proposed a deep neural network called DeepAds that can accurately localize the position of an advert in an outdoor scene image. This is particularly useful for advertisement and marketing agencies, for creating personalized video content for consumers. Our approach of localizing advert in a video frame is the first of its kind in the domain of advertisement and multimedia. We achieve the state-of-the-art results on a publicly available dataset of outdoor scenes. In the future, we will explore the possibility of using DeepAds model to identify new spaces in an outdoor scene, which can be used as viable locations for advertisement integration.

\bibliographystyle{IEEEbib}

\end{document}